\documentclass{article}

\usepackage{microtype}
\usepackage{graphicx}
\usepackage{subfigure}
\usepackage{booktabs} % for professional tables
\usepackage{natbib}
\usepackage{hyperref}
\usepackage{tcolorbox}

\usepackage{enumitem}

\usepackage[accepted]{icml2024}

\usepackage{amsmath}
\usepackage{amssymb}
\usepackage{mathtools}
\usepackage{amsthm}
\usepackage{multirow}

\usepackage[capitalize]{cleveref}
\usepackage[textsize=tiny]{todonotes}
\crefname{section}{Sec.}{Secs.}% change by chandan

\usepackage[labelfont=bf]{caption}
\usepackage{placeins}
\icmltitlerunning{Data Analysis in the Era of Generative AI}

\begin{document}

\newcommand{\chenglong}[1]{{\color{orange} \{chenglong: {#1}\}}}
\newcommand{\sdrucker}[1]{{\color{purple} \{sdrucker: {#1}\}}}
\newcommand{\vdibia}[1]{{\color{green} \{vdibia: {#1}\}}}

\newcommand{\ji}[1]{{\color{blue} \{jeevana: {#1}\}}}

\newtcolorbox{redbox}{
  colback=red!5!white,
  colframe=red!75!black,
  fonttitle=\bfseries,
  title=Challenges
}

\twocolumn[
\icmltitle{Data Analysis in the Era of Generative AI}

\begin{icmlauthorlist}
\icmlauthor{Jeevana Priya Inala}{}
\icmlauthor{Chenglong Wang}{}
\icmlauthor{Steven Drucker}{}
\icmlauthor{Gonzalo Ramos}{}
\icmlauthor{Victor Dibia}{}
\icmlauthor{Nathalie Riche}{}
\icmlauthor{Dave Brown}{}
\icmlauthor{Dan Marshall}{}
\icmlauthor{Jianfeng Gao}{}
\end{icmlauthorlist}
\icmlaffiliation{Microsoft Research}

\icmlcorrespondingauthor{Jeevana Inala}{jinala@microsoft.com}

\vskip 0.3in
]

\printAffiliationsAndNotice{} 

\begin{abstract}
This paper explores the potential of AI-powered tools to reshape data analysis, focusing on design considerations and challenges. We explore how the emergence of large language and multimodal models offers new opportunities to enhance various stages of data analysis workflow by translating high-level user intentions into executable code, charts, and insights.  We then examine human-centered design principles that facilitate intuitive interactions,  build user trust, and streamline the AI-assisted analysis workflow across multiple apps. Finally, we discuss the research challenges that impede the development of these AI-based systems such as enhancing model capabilities, evaluating and benchmarking, and understanding end-user needs.
\end{abstract}

\section{Introduction}
Data-driven decisions and insights are crucial across a wide range of industries and for everyday end-users. Businesses rely on data to make strategic decisions, healthcare providers track trends to improve patient outcomes, and journalists craft data-based news stories that inform the public. On a more personal level, individuals can use data to manage their finances, monitor their health, optimize their schedules, and even for shopping and making travel plans. However, the cost of current data analysis is high, meaning that only a select group of experts—data analysts—have the tools and skills to ask data-related questions, analyze information, and generate reports/insights based on these insights. For the rest of the population, this process is largely inaccessible; they must accept the insights provided by others, without the ability to ask their own questions or conduct data analysis for personal tasks.

Democratizing data analysis has the potential to revolutionize how we operate. Imagine a world where anyone, not just trained analysts, could effortlessly analyze data to make informed decisions. A small business owner could independently explore market trends, a patient could analyze their own health data to manage a chronic condition, or a traveler could optimize their trip based on past travel data. This shift would empower individuals and organizations, leading to more informed decisions.

\paragraph{\textbf{The Challenge of Data Analysis: A Complex and Iterative Process.}}
Deriving insights from data is a complex process. Data analysts need to iterate among task formulation, data collection, exploratory analysis, and creating visual representations to discover and validate data insights before drawing conclusions, and they further need to document the analysis as reports for sharing (\autoref{fig:ds-steps}). To achieve these diverse analysis steps, the analysts need not only conceptual knowledge (e.g., data sense-making, domain, statistics, visualization design knowledge) to make right analysis decisions, they also need tool expertise and programming skills to take actions. Furthermore, because the analysis process is rarely linear and different steps are segregated across various tools (e.g., once the analysts discover missing data based on their  current visualizations, they may need to backtrack and revisit data collection or cleaning steps), analysts face considerable overhead switching between tools and managing branching analysis steps. 

To address these challenges, many interactive and automated tools and solutions have been developed, both to enhance analysts' ability to understand and explore data and to close the gulf of execution. 
These include interactive visualization platforms like Tableau and Power BI, automated data preparation platforms like Alteryx or Trifacta, and programming environments like Jupyter Notebooks where data analysis can be integrated with data cleaning and visualization.
However, these systems often have to trade-off between the flexibility (and the expressiveness) of tasks that can be achieved with the system and the ease of learning and specifying the tasks. For example, 
compare Tableau, which provides user-friendly interface and drag-and-drop chart creation, with Python's Matplotlib library, which offers extensive customization and fine control over chart details but requires coding expertise.

\paragraph{\textbf{The Generative AI (GenAI) revolution}.}
The emergence of large language and multimodal models presents new opportunities to develop tools that are both expressive and user-friendly. These AI models possess the ability to perform human-like reasoning and transform high-level user specifications into low-level executable steps~\cite{bubeck2023sparks}. This reduces the user's need to learn new tools and languages while assisting with a variety of tasks.  Foundation models like ChatGPT,  GPT-4~\cite{achiam2023gpt}, Claude~\cite{anthropic2024claude}, Phi-3~\cite{abdin2024phi} have advanced proficiency in language understanding, conversation ability, and planning; they encode knowledge across various fields such as medicine and finance, along with code generation capabilities. Multimodal models like GPT-4o and Phi-3-Vision~\cite{abdin2024phi} further augment these abilities by integrating language and vision modalities, enhancing reasoning across diverse input forms.
With these abilities, we are in the era where AI-powered tools can empower novice users to perform data analysis tasks previously out of reach and even significantly boost the productivity of experienced data analysts.

\paragraph{\textbf{Uniqueness of AI-powered tools for data analysis.}} AI-powered tools are being developed for many domains, including video and image generation, gaming, medicine, and software engineering.
So, why does the data analysis domain require special treatment? Unlike generation tasks, data analysis involves planning and exploration over multiple steps. It also requires working with multiple modalities—natural language, structured data, code, images, and charts. Moreover, the task is fairly complicated that even non-AI solutions are spread across multiple apps and tools (for e.g.  a user might clean data in Excel, generate charts in PowerBI, and then create a report in PowerPoint). These complexities create interesting challenges for existing Generative AI models and systems, highlighting the need for better models and algorithms. Moreover, solving these tasks may require multiple AI tools and require novel techniques to avoid fragmented and disconnected experience for users. 

Data analysis is inherently iterative. Users must be involved throughout the process because the full specification of the task is usually unknown at the beginning. As users see initial results, they may want to follow up with new questions or adjust their goals. This iterative and multimodal nature of data analysis means that current AI tools, which often rely on  natural language interfaces and static modalities such as images and file upload, may not be well-suited for this domain. The data analysis domain inspires  invention of new interfaces and experiences that allow humans to collaborate with AI tools more effectively.

Moreover, data analysis is a sensitive domain. Errors can have serious consequences, especially in fields like healthcare or finance. Imagine an LLM that incorrectly identifies a trend in financial data, leading to poor investment decisions, or one that misinterprets medical data, resulting in incorrect diagnoses. This underscores the importance of accuracy and reliability in AI systems, as well as the need for infrastructure that allows end-users to verify and debug the outputs from these AI tools.

\paragraph{\textbf{Unlocking the potential of GenAI in data analysis.}}
The goal of this paper is 
to explore how we can  unlock the potential of generative AI tools in the data analysis domain. We aim to provide a view into parts of the vast design space of AI-powered data analysis tools that AI and HCI practitioners can use to organize existing AI systems/tools as well as identify areas requiring further attention. 
Our focus centers on three key themes.
\begin{itemize}
    \item \textbf{GenAI opportunities in data analysis.} We explore how GenAI systems can support each individual stage of the data analysis process. While the utilization of LLMs for code generation in tasks like data cleaning, transformation, and visualization is well-known, we delve into additional avenues such as aiding users in finding required data, ensuring analyses maintain statistical rigor, assisting in hypothesis exploration and refinement, and personalized report generation (see \autoref{sec:opportunities}).
    \item \textbf{Human-driven design considerations.} The design of these AI-powered tools and their user interfaces can have a huge impact on user experiences. For instance, some interfaces facilitate a more intuitive approach for users to specify intents such as selecting the color of a chart using a color picker widget vs. using natural language description. Similarly, distinguishing between seemingly correct but analytically different charts poses a challenge for users, especially without additional interface showing provenance analysis. Concretely, we discuss various design considerations for reducing users' intent specification efforts, enhancing users' ability to understand and verify system outputs, and streamlining the analysis workflow to reduce iteration overhead across multiple steps and tools (see \autoref{sec:design}). 
    \item \textbf{Research challenges ahead of us.} Finally, we conclude the paper by discussing the research challenges that must be addressed to implement AI-powered data analysis systems. These challenges include enhancing existing models' capabilities, addressing the scarcity of training and evaluation data, ensuring system reliability and stability of the data analysis process, and conducting user research to align system designs to meet users' cognitive abilities and practical needs (see \autoref{fig:challenges}).
\end{itemize}

\section{Background}\label{sec:background}

\subsection{The data analysis process}
\begin{figure}
    \centering
    \includegraphics[width=1\linewidth]{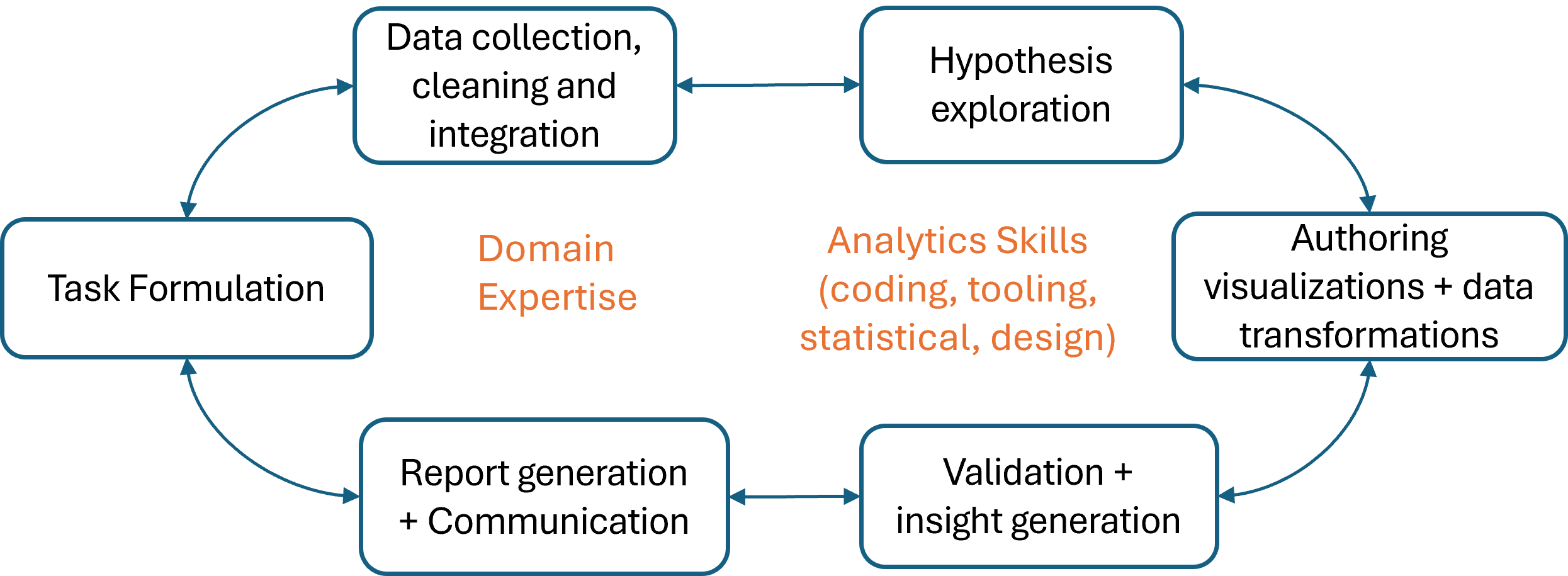}
    \caption{Overview of steps involved in deriving insights from data, highlighting the diverse skills required for these steps. We explicitly exclude any steps related to building/using models in this paper.  }
    \label{fig:ds-steps}
\end{figure}
\autoref{fig:ds-steps} and \autoref{tab:overview} show some of the common steps involved in deriving insights from data. This process has multiple steps and often involves iterating back and forth among these steps. The process usually starts with task formulation, where the problem, question, or decision to be addressed is identified  (e.g. What’s the trend of renewable energy adoption over the years for different countries?). This step also involves operationalizing the task into sub-tasks (such as figuring out global trends over time, ranking of countries, and compare trends among top countries). Then comes the task of collecting relevant data to answer the above questions. This might involve actions like querying databases, instrumenting applications, web scraping, data cleaning, and integration from multiple sources. The sequence of these steps can vary; analysts may already possess data and seek insights or may identify data gaps or data cleaning needs later in the analysis.  

Following data collection, the exploration phase begins, where analysts delve into the dataset to gain familiarity and identify potential patterns or relationships. This involves forming further sub-tasks such as generating descriptive statistics (e.g., mean, median, and count), visualizing the relationship between several variables, trends, and correlations, and generating hypotheses to explore. Then, they transform data or generate visualization artifacts to help support their hypotheses.
Once artifacts (transformed data and charts) are generated, one has to understand them and validate them to ensure the reliability and accuracy of their findings. This may involve statistical analysis, domain knowledge-based validation, or comparison with external sources. The insights gathered from these artifacts are then used to refine the hypotheses accordingly. 

Finally, analysts communicate their findings and generate the final results. This may involve creating reports, dashboards, or presentations to effectively communicate insights to stakeholders. Analysts may iterate on their communication based on the feedback received, further refining their presentation of the data and insights.

The data analysis process is inherently complex and multifaceted, characterized by its iterative nature and the diverse set of skills required at each stage. Each step demands a distinct set of domain knowledge, statistical expertise, coding proficiency, and familiarity with various tools. Moreover, the fragmented nature of the process requires analysts to continuously juggle between different tools and methodologies to derive meaningful insights. 

Please note that the steps outlined above are not exhaustive. They do not cover areas such as model building, automated machine learning, collaborative authoring, or accessibility. While these topics are crucial to data science and analysis, they require specialized attention beyond the scope of this discussion. Instead, we focus on the iterative and exploratory aspects of deriving insights from data and how generative AI can influence this process.

\subsection{LLMs, LMMs, and agents} 
The research on LLMs, LMMs, and the AI systems that are built using these models has advanced drastically in recent years. 
On the base level, we have several foundation models starting with language models (ChatGPT,  GPT4~\cite{achiam2023gpt}, Claude~\cite{anthropic2024claude}, and  Llama~\cite{touvron2023llama}) and then recently multimodal models (GPT-4o~\cite{gpt4o}, LLava~\cite{liu2024visual}, and Phi-3-Vision~\cite{abdin2024phi}). These models are well-known for their ability to generate and understand code and have been shown to have good performance in generating algorithms and solutions to interview/competition problems~\cite{li2022competition}, write real-world software with GitHub copilot~\cite{nguyen2022empirical}, and various data transformations and visualization authoring tasks~\cite{dibia2023lida, narayan2022can}.  These models are also instruction-tuned to understand and converse with people in a chat-like environment~\cite{ouyang2022training}. 

Using foundation models as a basis, researchers have developed AI agents or AI assistants tailored to different domains. These agents act as interfaces around foundation models, embedding any application-specific knowledge through system prompts or leveraging external tools like code execution engines, search engines, or calculators to respond to user queries~\cite{lu2024chameleon}. For example, a code interpreter~\cite{codeinterpreter} is an AI assistant that can generate code, execute it in a sandbox environment, and use the execution results to produce further text or code.

On top of all of these, there are also multi-agent AI systems that have a series of agents, each supposed to be specialized in a particular subtask, with all agents acting together to collaborate and respond to a user query~\cite{wu2023autogen, hong2023metagpt}. For example, one such multi-agent system is a generator and a evaluator-agent combination that is  used to enhance the reliability of AI systems~\cite{du2023improving}. 

To establish terminology, in the rest of the paper we use ``AI systems'' to refer to systems powered by the latest breakthroughs in LLMs, LMMs and Generative AI, rather than traditional symbolic AI systems. In some places, we also use LLMs and LMMs explicitly based on the context.

\section{Case study: User experiences with AI-tools for visualization authoring} \label{sec:example}
\begin{figure*}
\centering
\includegraphics[width=0.65\linewidth]{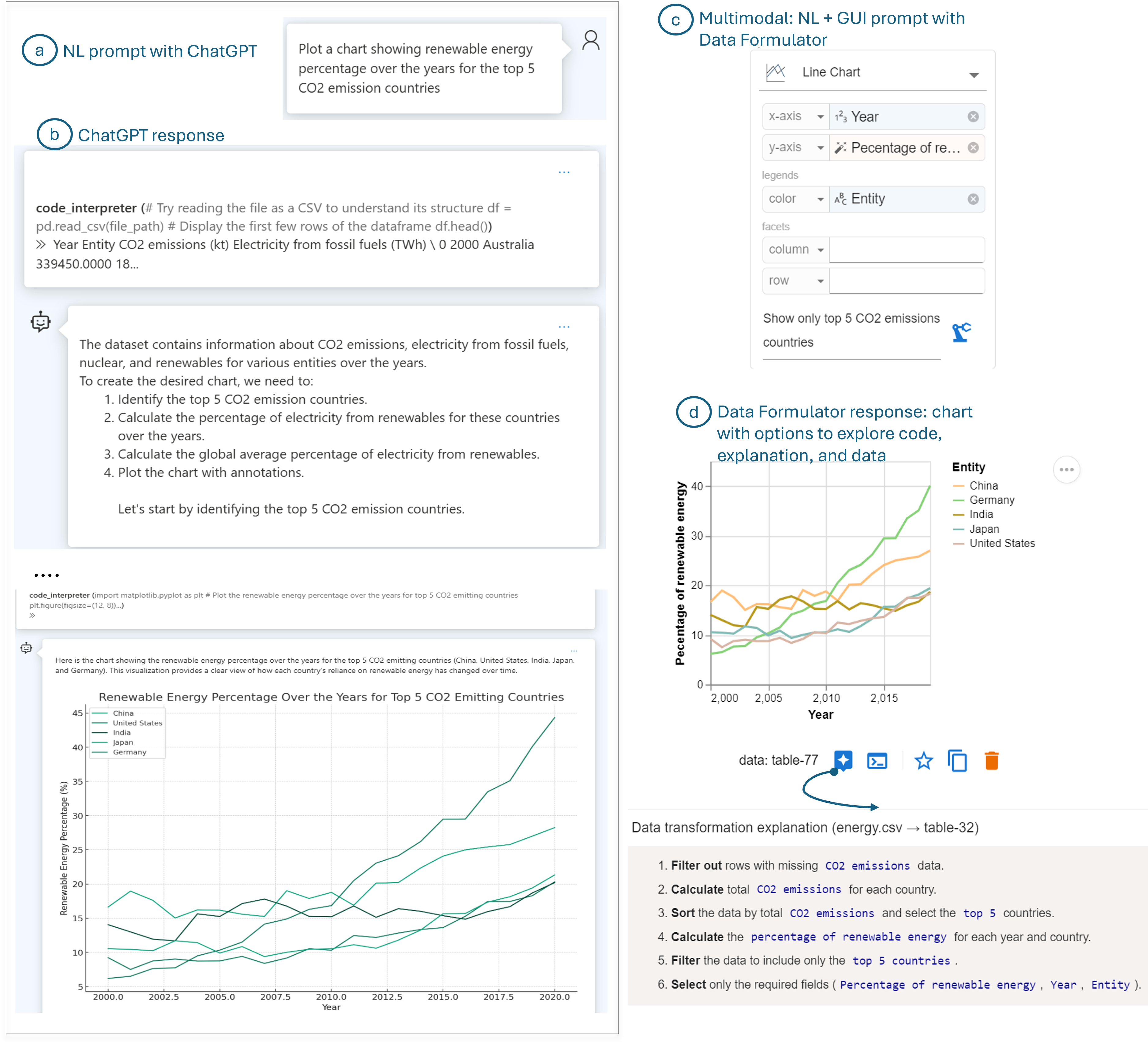}
\caption{User experience for providing inputs and output formats with ChatGPT vs Data Formulator~\cite{wang2023data}}
\label{fig:run_ex_io}
\end{figure*}

Data visualizations are prevalent in different data analysis stages: data analysts create visualizations to assess data quality issues, explore relations between data fields, understand data trends and statistical attributes, and communicate their insights from data to the audience. Although modern visualization tools and libraries have greatly eased the requirements of user expertise and effort, the authoring process remains challenging since the analysts not only need to learn to use visualization tools to implement charts, they also need to transform data into the right format for their chart design. 

With the emergence of LLMs, AI-powered visualization authoring tools have been developed to reduce the authoring barrier~\cite{dibia2023lida, wang2023data, maddigan2023chat2vis, vaithilingam2024dynavis}. Here, we use an example visualization task to compare user experiences with (1) traditional programming tools, (2) direct conversational interaction with LLMs via a chat-based interface, and (3) LLM-powered interactive data-analysis tools, highlighting various human-driven design principles which will be discussed later in this paper. We center our comparison around the user experience for (1) specifying the initial intent, (2) consuming the AI system's output, (3) editing, (4) iterating, (5) verifying AI system's output, and (6) the workflow in the larger data analysis context. 
\begin{table}
    \centering
    \includegraphics[width=\linewidth]{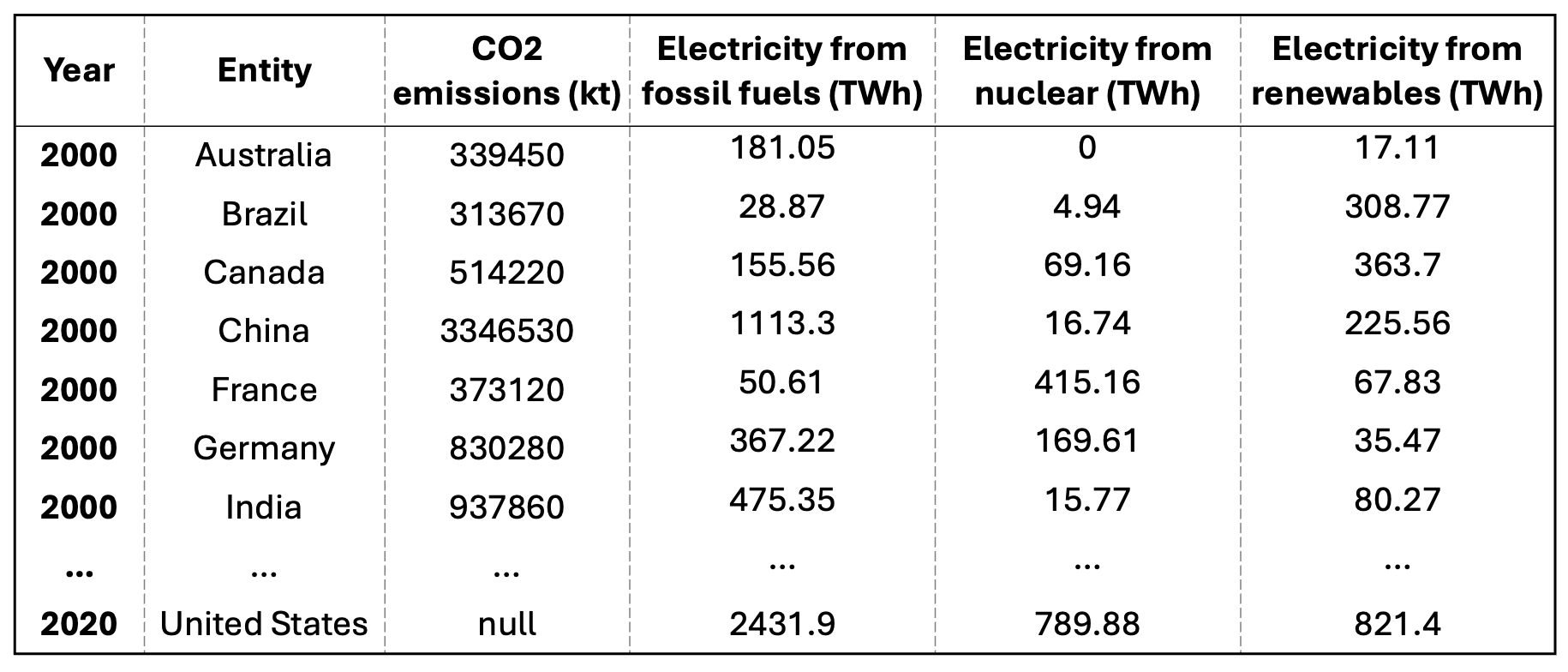}
    \caption{Example dataset: CO$_2$ and electricity data of 20 countries between 2000 and 2020.  Target visualization: Renewable energy percentage over the years for the top 5 CO2 emission countries. Dataset obtained from \cite{dataformulator2}. }
    \label{tab:global-energy-datatable}
\end{table}

\subsection{Example task and the traditional experience} 
\autoref{tab:global-energy-datatable} shows a sample dataset that shows the CO$_2$ and the electricity generated from different sources for 20 countries between the years 2000 and 2020. Let us assume the user is at a stage in the data analysis pipeline where they want to investigate the trend of  percentage of electricity from renewable energy sources over time for the five countries with the highest CO$_2$ emissions.

The traditional (non-AI) approach to this task involves two steps: (a) data transformation, where the user creates new columns (e.g., "percentage of renewable electricity") and performs operations like grouping, summing, and filtering. More complex tasks may require pivoting, unpivoting, and other complex data transformations; and (b) chart authoring, where the user determines the chart type and encodings to appropriately represent the data trends. Data transformations can be performed using various methods, such as formulas in Excel or Python libraries like Pandas within a Jupyter notebook environment. Similarly, charts can be created with libraries such as Seaborn and Matplotlib or in interactive environments like PowerBI and Tableau. While this traditional approach offers full control for users, it presents a steep learning curve for beginner data analysts and non-programming end-users. 
\begin{figure}
\centering
\includegraphics[width=0.99\linewidth]{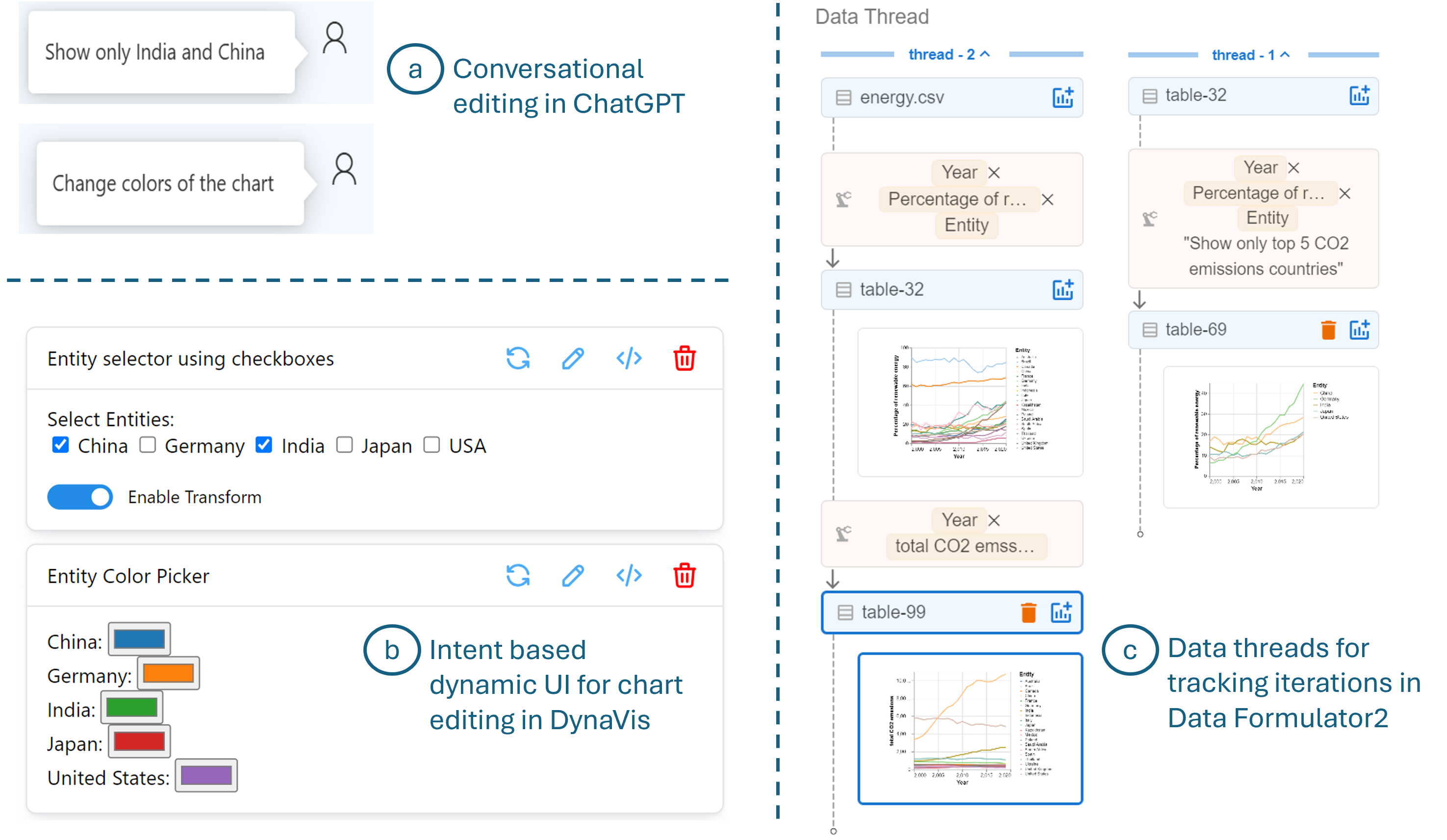}
\caption{Constrasting conversational editing and iterations in (a) ChatGPT with (b) dynamic intent based UIs in DynaVis~\cite{vaithilingam2024dynavis}  and (c) data threads in Data Formulator2~\cite{dataformulator2}.}
\label{fig:run_ex_it}
\end{figure}

\subsection{Conversational LLM interface (ChatGPT with GPT-4o and CodeInterpreter)}
LLMs have made the above process easier for a wide range of users with different experience levels. For this running example, we use ChatGPT's interface with GPT-4o model as the conversational LLM together with CodeInterpreter assistance capability. In this conversational experience, the user loads the dataset and expresses their intent as a natural language command, as shown in \autoref{fig:run_ex_io}(a). ChatGPT processes the request and performs several steps such as loading and understanding the dataset, figuring out subtasks, writing code for the subtasks, and executing the code to generate the final chart. ChatGPT shows its steps as code snippets, natural language descriptions, and renders the final chart for the user (\autoref{fig:run_ex_io}(b)). For this 
 task, ChatGPT requires multiple invocations of the underlying model for the different sub-tasks, leading to long wait times for users. Moreover, repeated model invocations increase the likelihood of failure, and the system can sometimes get stuck on one of the steps, making it difficult to recover.

In conversational LLM interface, users can edit and iterate on their visualizations through follow-up instructions in a conversational manner (see \autoref{fig:run_ex_it} (a)). The chat interface also allows users to backtrack to a previous conversation and edit from there. However, there is no easy way for users to track the different iterations. The ChatGPT system might also lose some context (such as an already performed data transformation) while backtracking, leading to the need for re-computation. Since the non-deterministic nature of the underlying model can produce varying outcomes upon re-computation, this again increases the chances of failure and leads to unreliable user experiences. 
\begin{figure*}
\centering
\includegraphics[width=0.7\linewidth]{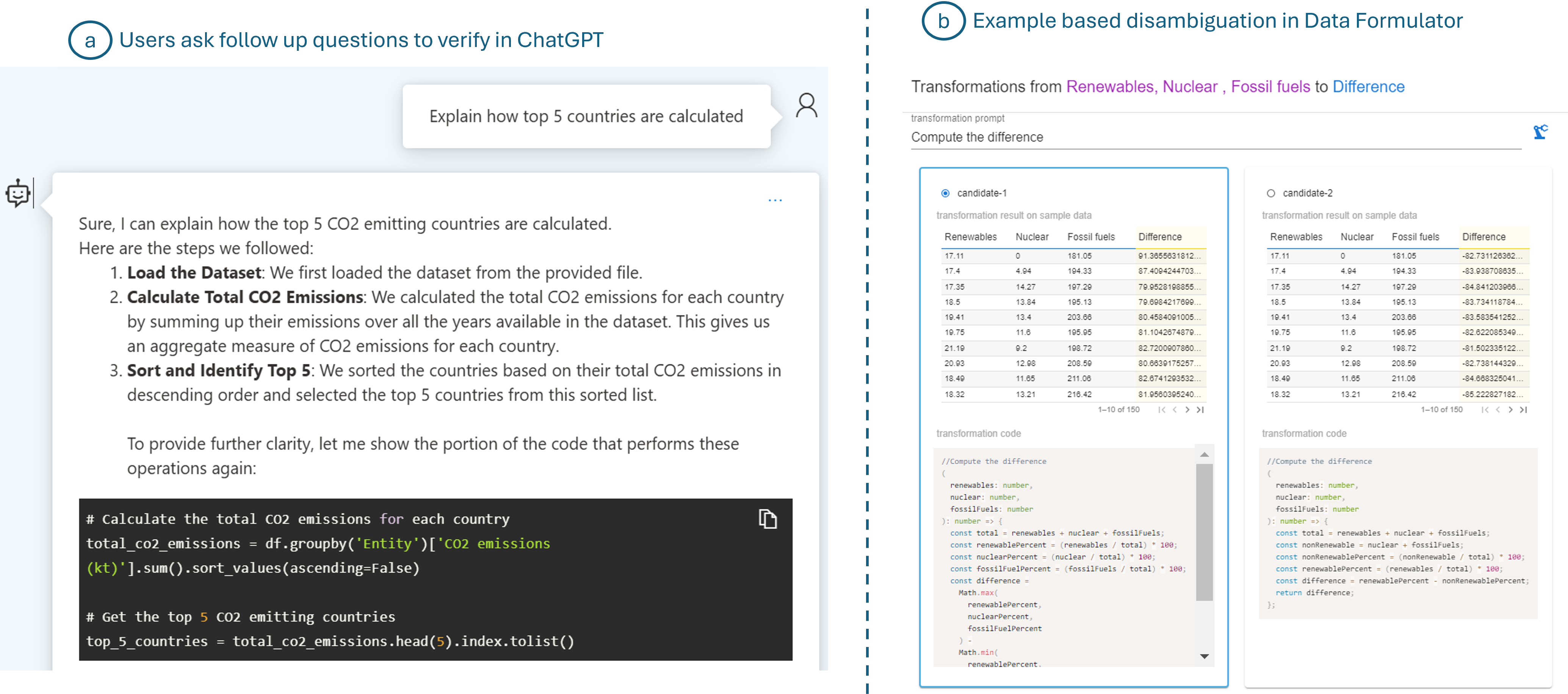}
\caption{User experience for trust and verification of AI outputs in ChatGPT vs Data Formulator.}
\label{fig:run_ex_trust}
\end{figure*}

The latest ChatGPT interfaces and similar systems like Claude~\cite{anthropic2024claude} support multimodal prompts, including natural language, images, and other documents, yet the intention remains primarily mediated through natural language, which can limit expressiveness.  For example, some intents, like the exact color preferences in RGB values, are difficult to specify in natural language but easy to select from a color picker. Other commands can be imprecise, such as ``put that there'' in a complex figure, where users could more accurately specify the context by directly selecting the region. Additionally, users might want to repeat certain actions multiple times with slight variations to see their impact  (for e.g. try out different colors before choosing one), but with conversational LMMs, they have to invoke the model for each change, which can again lead to different outcomes each time. 

 The conversational framework also allows users to ask for further clarification or explanation of how the model performed a certain computation in its analysis (\autoref{fig:run_ex_trust} (a)). The system responds back with the explanation as well as code snippets to help users understand and make any changes if needed, allowing users to trust and validate the model's response. 

 Other examples of conversational AI interfaces include the copilots or AI assistants for individual tools such as Microsoft's Office Copilot. However, these copilots function as separate entities. Consequently, if a user needs to perform a task using multiple apps (e.g., using Excel to clean the data and PowerBI to create charts and dashboards), they must manually switch and transport the context between these tools and their respective copilots (\autoref{fig:run_ex_apps} (a)), which leads to an inefficient workflow. 

\subsection{LLM-powered interactive data analysis tools}\label{sec:runex1}
Now, we describe a different user experience for the above case study using a few selected interactive LLM-powered data analysis tools.  Some systems and experiences rely on the capabilities of LLMs to support data analysis, while departing from the types of chat-like experience described above.
First, for specifying the user's intent, Data Formulator~\cite{wang2023data} has devised a multi-modal UI that allows users to specify the intent using a combination of field-encodings and natural language (NL) instructions (\autoref{fig:run_ex_io}(c)).  Based on the user's multi-modal inputs, an LLM model generates the necessary data transformations to create a new table. Using this table and the user's specified chart encodings, a Vega-Lite specification can then be straightforwardly generated to produce the chart. The chart is then rendered along with options for users to read/edit code, read code explanations, and view the newly transformed data table (\autoref{fig:run_ex_io} (d)).
This multi-model UI allows users to have more control on the type/orientation/high-level details of the chart they want to create but also offload low-level tasks, such as performing the necessary transformations needed for a new concept, to the LLM. 
From the multi-modal input, the model also has more information to ground the user’s instruction for better code generation. For instance, the system can generate a Vega-Lite specification without requiring a separate model invocation.
Moreover, this AI system is tailored for data visualization tasks, incorporating components for data processing and summarization, along with domain-specific prompts and instructions for LLMs.  These features enhance reliability and reduce latency in the AI-driven data visualization authoring process compared to the experience with the ChatGPT interface.

DynaVis~\cite{vaithilingam2024dynavis} is another interactive AI tool that provides a multi-modal UI for users to specify their chart editing intents. Based on the high-level intent specified in natural language, such as ``I want to change the colors of this chart,'' the system dynamically generates widgets (e.g., color pickers) on the fly, allowing users to try out several similar options (see \autoref{fig:run_ex_it}(b)). The system only calls the LLM when initially generating the widget, not for every change requested by the widget, enabling users to experiment with many variants and receive instant feedback for their edits. These intent-based widgets help users specify intents that are difficult to express in natural language and give users a sense of control and trust by allowing them to easily explore various chart editing options.

Data Formulator2~\cite{dataformulator2} extends Data Formulator~\cite{wang2023data} by supporting iterative exploration as a first-class interaction. It allows users to pick any existing visualization and express their modification intent again through the multi-modal UI from \autoref{fig:run_ex_io} (c). The system's agents reuse previously generated code and data to generate new results, thus giving a consistent experience to users. The system organizes the user's interaction history as data threads (see \autoref{fig:run_ex_it} (c)) to help the user manage the analysis session. Data threads enable users to easily locate existing plots for refinement, branch out from previous steps to explore alternatives, or backtrack to correct mistakes.

As a form of trust and verification mechanism for users, Data Formulator creates multiple chart/transformation candidates based on the user's intent. It allows users to inspect these candidates via their code and a few examples (see \autoref{fig:run_ex_trust}(b)), enabling users to disambiguate or refine their specifications. 
\begin{figure*}
\centering
\includegraphics[width=0.7\linewidth]{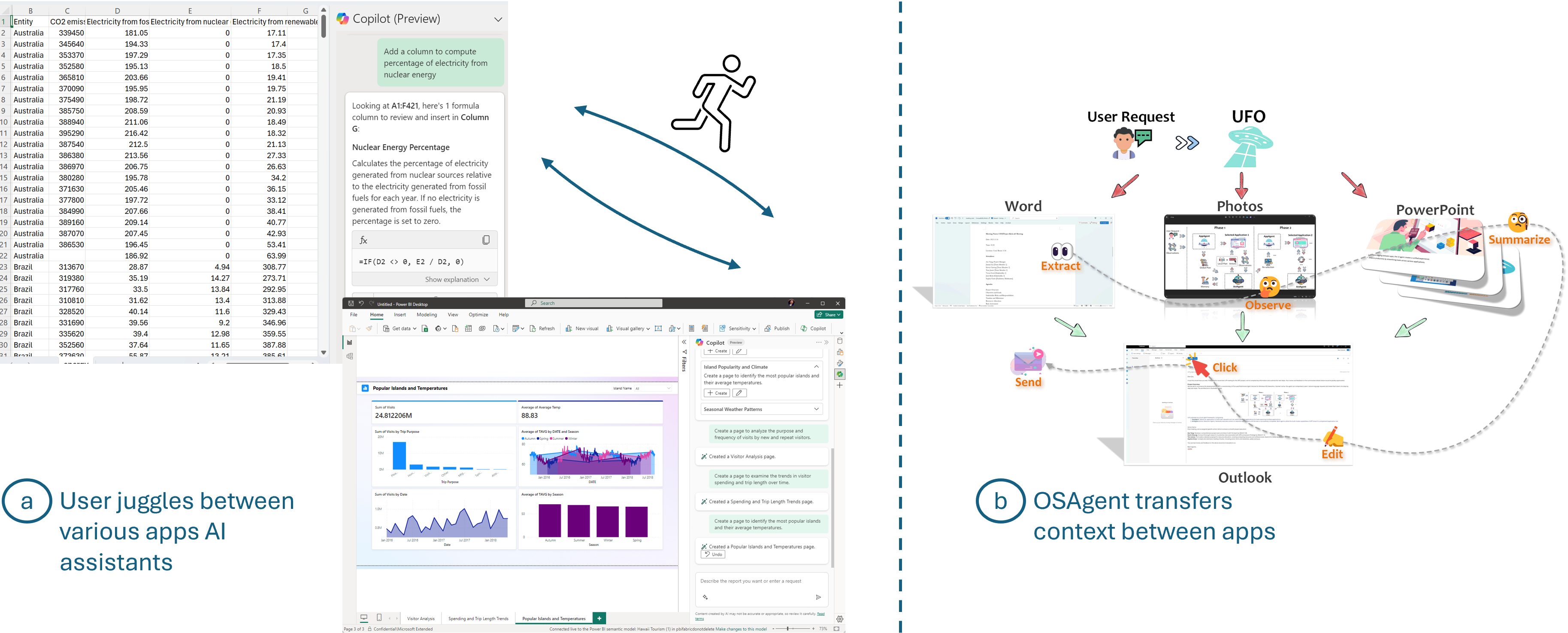}
\caption{User experience for multi-app workflows with (a) existing  AI assistants for individual apps vs (b) an OSAgent~\cite{zhang2024ufo} that transfers context between apps. Figure obtained from~\cite{zhang2024ufo}.  }
\label{fig:run_ex_apps}
\end{figure*}

Finally, to streamline the data analysis workflow, Data Formulator combines both data transformations and visualization authoring into a single tool, eliminating the need to switch between separate tools for each step. Further streamlining the workflow are multi-app systems such as UFO~\cite{zhang2024ufo}, which is an innovative UI-focused agent tailored for applications on Windows OS using GPT-Vision (\autoref{fig:run_ex_apps}(b)). UFO employs models to observe and analyze GUI and control information, allowing seamless navigation and operation within and across applications, simplifying complex tasks.

\subsection{Remarks} In our comparative analysis of user experiences with various AI tools for visualization authoring, we observe that the design of these tools significantly impacts users' ability to create the desired visualizations. These design considerations are critical not only for  visualization authoring tools but also for AI tools across the different stages of data analysis. In \autoref{sec:design}, we elaborate on these design principles and explore various strategies--beyond those examined in this case study--to alleviate user specification burdens, enhance trust and verification strategies, and facilitate workflows across multiple stages and tools.

Before delving into these design principles, \autoref{sec:opportunities} will investigate additional AI opportunities within the broader data analysis landscape, extending beyond visualization authoring.

\section{Opportunities for AI systems in the data analysis domain}\label{sec:opportunities}

\begin{table*}
\centering
    \begin{tabular}{|p{2.7cm}|p{5.5cm}|p{7.0cm}|}
    \hline
    \textbf{Step} & \textbf{Description} & \textbf{GenAI opportunities}  \\
    \hline
         Task formulation & Identify the problem to be solved, the question to be answered, or the decision to be made.& Helping people go from fuzzy specifications to concrete, measurable tasks; Adding domain knowledge to task formulation; Finding similar tasks and analyses. \\
         \hline
         Data collection, cleaning, and integration & Collect relevant structured data and prepare it for analysis. & Finding appropriate data from existing sources; Extracting data from unstructured documents or media; Executing extraction code from relevant databases/web; Automated and semi-automated cleaning or anomaly detection and integrating data from multiple sources.  \\
         \hline
         Hypothesis exploration & Generate hypotheses about data to understand the data. & Domain knowledge based exploratory analysis; appropriate statistical tests based on task and data.\\
         \hline
         Execution and authoring & Create visualizations, tables, or other forms of structured data representation to support the hypothesis. & Circumventing deep procedural knowledge required for use of libraries and tools (e.g., Pandas, PowerData, etc.). \\
         \hline
         Validation and insight generation & Read charts and data to validate hypothesis, generation, and assumptions; gather insights for downstream analysis. & Automatic evaluation and repair of analysis; validate analysis and assumptions with domain knowledge; multi-modal analysis to generate insights from data and charts, interpretable results. \\
         \hline
         Report generation and communication & Make decisions/choose actions and communicate results to stakeholders. & Interactive decision support, automatic recommendations based on analyses; customized representations for different presentations and audiences; dashboard generation; generating creative charts and infographics.\\
         \hline
         \hline
         Domain and analytical skills  & Need coding, domain knowledge, statistics, and tool skills for different stages of the analysis pipeline. & Low/no code experiences; Offer domain knowledge support and statistical expertise; tool copilots \\ 
         \hline
    \end{tabular}
    \caption{Various analysis steps along with potential opportunities for assistance that GenAI can provide.}
    \label{tab:overview}
\end{table*}    

In this section, we explore how AI systems can help users navigate the complexities of data analysis workflows. Each stage of the analysis process presents unique challenges, from complex reasoning tasks and iterative processes to the need for diverse skills in coding, domain knowledge, and scientific methodologies across multiple tools. To address these challenges, AI systems offer numerous opportunities to streamline and enhance these workflows.

The goal of this section is to establish some concrete problem definitions for a few potential AI-driven workflows in the data-analysis domain (i.e. establishing user's requirements and  expectations from the AI system). While there is an aspiration to automate the entire analysis process, it is imperative to dissect the process into its sub-tasks to identify areas where AI systems can make a significant impact and where they may fall short. This way of solving a complex problem by breaking into multiple sub-tasks also helps users evaluate the intermediate outputs of the AI systems and to intervene and iterate when necessary, thus helping to ensure the reliability and effectiveness of the overall process.

Once we establish the opportunity and its impact on users, we do a literature review of studies that are related or useful for that particular opportunity. 
This is not an exhaustive set of opportunities or related works, but rather an preliminary overview aimed at establishing a foundation on which future researchers can build upon.

\subsection{Closing the skills gap: Empowering users in data analysis}\label{sec:common-opportunities}
A data analysis task can require diverse skills including domain knowledge, data science knowledge, coding, as well as knowledge of and how to use tools. The need to possess this diverse skill set often acts as a barrier for non-expert individuals, hindering their ability to engage in effective data analysis and fully realize its benefits. LLMs bring new opportunities to mitigate this barrier by supplementing the existing expertise of users with complementary skills. 
\paragraph{Low code/no code experiences.}
During a data analysis process, once users identify a specific next step, such as visualizing temporal trends, they still face the additional challenge of writing code or using specialized tools to accomplish this objective. For non-programmers, LLMs can alleviate this burden by aiding users in seamlessly transitioning from intent to automatically generating the required code.

LLMs have shown great proficiency in generating code snippets for data analysis from natural language descriptions, especially in languages such as Python. There are several AI systems for taking a user's specification in natural language or other forms and generating code for various sub-stages such as data cleaning, transformations, querying databases, and visualization authoring.

LIDA \cite{dibia2023lida} is an open-source tool that enables goal-based data visualization generation by generating, refining, filtering, and executing code to produce visualizations. It also includes an infographer module that creates data-faithful, stylized graphics using image generation models. The system supports multiple Python libraries, including Matplotlib, Seaborn, and Altair, for visualization generation.
ChatGPT's code interpreter API~\cite{cheng2023gpt} lets users upload their dataset and pose queries in natural language (e.g., analyze the trend of renewable energy over time). Then the model generates a code snippet for this task and executes it to generate a plot. Chat2Vis~\cite{maddigan2023chat2vis} and ChartGPT \cite{tian2024chartgpt} are some of the other LLM-based tools that generate code snippets for data analysis and visualization tasks based on natural language input. 
DataFormulator~\cite{wang2023data}, which was introduced in \autoref{sec:runex1}, allows users to specify their visualization intent through drag and drop in addition to natural language and uses these multi-modal contexts to automatically generate the code for visualization. Some of these tools also allow users who may be slightly more proficient in coding skills to inspect the generated code and make any changes if needed.

There are also systems to enable processing of datasets~\cite{sharma2023automatic}, performing data cleaning~\cite{singh2023datavinci, narayan2022can} and transformations~\cite{joshi2023flame} by incorporating domain knowledge in natural language form.
Another well-established area of research, known as semantic parsing, involves converting natural language to SQL queries~\cite{li2024can, Floratou2024NL2SQLIA} for extracting data from databases.

However, the effectiveness of these systems may differ, highlighting the need to improve reliability and the need to establish evaluation benchmarks and metrics. Additionally, these tools can benefit further from the design considerations we present later, such as enabling natural forms of interactions and fostering user trust in the process.

\paragraph{Statistical assistance.} 
While AI systems for coding components of the data analysis pipeline have received attention, the exploration of these models for other skills, such as statistical proficiency, remains relatively limited. 

Here are a few examples where LLMs can offer statistical assistance: LLMs can help users select appropriate statistical tests based on tasks and data, such as t tests to compare the means of two groups. They can also help users avoid common pitfalls~\cite{zuur2010protocol} such as selection bias, misinterpretation of p-values, and overfitting by providing guidance on issues like multiple comparisons and confounding variables. 

AI systems that expose LLM's statistical reasoning capabilities for the above cases and trigger them at relevant stages during the analysis process hold significant promise.
Some relevant prior work in this area comes from the pre-LLM era. Systems such as~\cite{jun2019tea, jun2022tisane} are steps towards simplifying statistical analysis by allowing users to provide domain knowledge and automatically extracting statistical constraints. 
%\ji{Any other citations? }

\paragraph{Domain knowledge support.}
LLMs contain vast amount of knowledge about various domains from their pre-training datasets which can enable us to automate domain-specific contextual understanding of the data and the task~\cite{wadden-etal-2022-multivers, liang2023gpt4}. For example, LLMs can help users understand the various columns in a dataset, help generate semantically rich features for downstream analysis tasks~\cite{hollmann2024large}, and can help users interpret the results (such as finding anomalies~\cite{lengerich2023llms}) of the analysis within the domain context.

\paragraph{Tool copilots.}
Recently, numerous applications feature their own AI copilots or AI assistant systems to facilitate a natural language and chat-based interface for their tools. Examples include BingChat, Office365 copilot, PowerBI copilot, Einstein copilot for Tableau and so on. These copilots notably lower the entry barrier for users engaging in data analysis tasks with their tools. In some instances, these copilots also act as an intelligent tutor that guides users to familiarize themselves with complex software interfaces, offering step-by-step instructions and clarifications on tool functionality.

Currently, these tools are application specific. Each app designs their own API/domain-specific language and uses an LLM/LMM (such as GPT4, Gemini, Llava)  to trigger appropriate APIs based on the user's chat context. The language model's ability to follow instructions and APIs is a crucial component here. To reduce the burden on copilot developers, AI systems that can automatically understand and utilize existing tools and APIs offer a promising solution, extending AI assistance to a broader range of applications.
In \autoref{sec:multi-app}, we also discuss design considerations for enabling these copilot frameworks  to communicate with each other to provide a more streamlined experience for users. 

In summary, LLMs/LMMs exhibit proficiency in coding, statistical analysis,  domain knowledge, and tool usage, opening up opportunities to not only enhance
users' data analysis abilities but also lower the skill barrier of entry to do so. 

\subsection{Potential AI workflows for the different stages of data analysis process}
\paragraph{Task formulation.}
One of the first steps of the analysis process is to formulate high-level goals for the analysis task and  refine them into specific actionable tasks.
Often, people face a cold-start problem, where they are unsure of what kind of analysis to perform or what insights they want to gain from their dataset. In such cases, AI systems can assist by leveraging domain-knowledge to understand the dataset and formulate meaningful questions. LIDA’s goal explorer module~\cite{dibia2023lida} addresses this problem by generating high-level goals grounded in data and an optional persona. Each hypothesis is generated as a triplet - goal (e.g., Understand trends in $CO_2$) a visualization that addresses the hypothesis (e.g., Line chart of $CO_2$ per year) and a rationale on why the goal is informative.

LLMs can also assist in translating high-level user requirements (e.g., evaluating a company's performance) into specific and actionable tasks (e.g., assessing revenue, employee satisfaction, etc.) again by leveraging the domain knowledge.
Some AI systems to refine fuzzy specifications include InsightPilot~\cite{ma2023insightpilot} takes a vague user specification and uses an LLM to issue concrete analysis actions that are then given to a dedicated insight engine to explore data and generate insights. NL4DV Toolkit~\cite{narechania2020nl4dv} is another tool that   generates analytic specifications for data visualization from natural language queries by breaking down an input query into attributes, tasks, and visualizations.

In addition to assisting with goal formulation and refining user tasks, AI systems can further enhance the analysis process by drawing inspiration from existing examples.
People sometimes get inspiration from existing charts and data analysis~\cite{bako2022understanding}. Similarly, AI systems can also assist by identifying similar datasets and questions, and their data analysis tasks and use them to guide the user into formulating their own task for their dataset. This draws some similarities to Retrieval Augmented Generation (RAG) based approaches that are commonly used to overcome a model's limited prompt window size by selecting relevant sources from a vast body of external knowledge~\cite{lewis2020retrieval,izacard2022atlas, peng2023check, lu2024chameleon} and also few-shot learning approaches with examples chosen based on a similarity metric~\cite{poesia2022synchromesh, li2023unified, liu2021makes}. 

\paragraph{Hypothesis exploration.}
Another avenue for AI systems to assist is by reducing the risk of biased data analysis through the support for multiverse analysis, wherein various analytical approaches are systematically examined. This helps users evaluate the strength of conclusions and improves transparency in decision-making by presenting all potential outcomes.
Boba~\cite{liu2020boba} is an early pre-LLM system to simplify multiverse analysis in data science by letting analysts write shared code with local variations, generating multiple analysis paths.  

Automated hypothesis exploration, while highly beneficial, poses significant challenges due to the need to mitigate exponential blow-up of hypothesis space.  Hence, it is an interesting research opportunity for AI systems to carefully combine domain knowledge and iterative exploration and at the same time avoid LLM's own inherent biases. 

\paragraph{Execution + Authoring.} Given a concrete goal, the next step is to generate the code necessary for any data transformations and for configuring the charts for the visualization. There are numerous works on automating this process with LLMs and we refer to ~\cite{dibia2023lida, wang2023data, maddigan2023chat2vis, shen2022towards} for a more thorough exploration of these works.

\paragraph{Validation, insight generation, and  refinement.}\label{sec:refine}
Once data analysis artifacts (charts and new data representations) are created, it is now time to inspect them carefully. This inspection serves two primary purposes. The first is to validate the analysis and assumptions against general expectations derived from domain knowledge. The second is to extract insights that can inform downstream iterative analysis and decision-making processes.
Vision language models such as GPT-V and Phi-3-Vision, have shown a lot of potential in analyzing images, charts, and figures. This gives us the opportunity to use these models to reason about the visualizations and generate insights from them.
     
LLMs and LMMs have been used to evaluate visualizations (or their representations). LIDA~\cite{dibia2023lida} uses an LLM to examine the generated code for a visualization and judge it based on a set of predefined criteria. ChartQA~\cite{masry2022chartqa}, PlotQA~\cite{methani2020plotqa} datasets are established to test LMM performance in answering charts-based questions. There are several models~\cite{han2023chartllama, masry2023unichart} and approaches~\cite{obeid2020chart, huang2024pixels} in the literature for using LLMs and LMMs to understand and gather insights from charts. LLMs have also been instrumental in consolidating insights from various analysis steps to construct a cohesive and comprehensive narrative~\cite{zhao2021chartstory, lin2023inksight, weng2024insightlens}.

Finally,  these chart understanding capabilities of vision models can be used to automatically iterate and refine hypotheses. For example, for regression analysis, the AI system can fit various curves to the scatter plot of the data, analyze the goodness of fit for each, and try other types of curves based on the initial analysis.   This also draws similarity to LLM-based self-repairing for code~\cite{olausson2023self, le2023codechain} works where LLMs reason about their own generated code and refine them further. 

It is important to note that the user might still play an instrumental role in validation of the AI assistant's output. Gu et al.,~\cite{gu2024analysts} found that providing a combination of a high-level explanation of the steps that the AI is taking, the actual code, as well as intermediate data artifacts is important to help resolve ambiguous statements, AI misunderstandings of the data, conceptual errors, or simply erroneous AI generated code. We discuss more on user-verification strategies in~\autoref{sec:trust}.

\paragraph{Data discovery.}  
In many data analysis workflows that leverage AI, it is often assumed that users have access to all the necessary data. However, ongoing data analysis may reveal the need for additional data to enhance the depth and accuracy of the analysis. For example, an analysis focusing on county-based statistics may require population data to perform proper averaging. Alternatively, some users may prefer a streamlined experience akin to a search engine, where they can pose direct questions like "which car should I buy?" instead of searching for the required dataset themselves. AI systems present an opportunity to help users locate the necessary datasets in such situations~\cite{fernandez2023large}. 

Data discovery in a data lake is a well-known problem. LLMs extend beyond traditional keyword/tag -based search methods, offering natural language query-based search in data lakes. By learning to embed popular datasets in a semantically meaningful form, LLMs facilitate dataset retrieval based on user queries.
For example, Starmie~\cite{fan2022semantics} uses pre-trained language models like BERT for semantic-aware dataset discovery from data lakes, focusing on table union search.

Furthermore, AI systems are capable of interacting with various web APIs, such as flight APIs and web search APIs, to dynamically extract data~\cite{liang2023taskmatrix}. For example, WebGPT~\cite{nakano2021webgpt} can browse the Web to answer long-form questions. There are other AI systems that extract real-world knowledge using a search engine to answer questions~\cite{lu2024chameleon, schick2024toolformer, peng2023check}. This opens up possibilities for LLMs to gather data from diverse websites based on user queries, organize the collected information into structured tables, and utilize it for reasoning and analysis. Although there have been previous non-AI based web automation tools~\cite{barman2016ringer, inala2017webrelate}, LLM-based approaches offer enhanced capabilities in data extraction and knowledge synthesis.

AI systems can also help with data cleaning and integrating multiple datasets from different sources~\cite{narayan2022can}. This can include handling data formatting issues, resolving inconsistencies, and merging datasets to create a unified dataset for analysis. 
   
Another significant opportunity lies in extracting structured data from non-tabular data formats like customer reviews, meeting notes, and video transcripts. LLMs can address this by analyzing the text data to extract structured data such as sentiments and trends for downstream analysis.
There are several recent works on using LLMs to convert unstructured data to structured data~\cite{vijayan2023prompt}, particularly in the medical domain~\cite{pmlr-v225-goel23a, bisercic2023interpretable}. Another example is HeyMarvin~\cite{heymarvin}, a qualitative data analysis platform, that exemplifies this approach by (semi) automating tasks such as transcription and thematic coding of user interviews, facilitating deeper insights into customer needs and preferences. However, there is still room to further improve this process by integrating it into the broader task context, enabling the extraction of necessary features based on the specific requirements of the data analysis task.

\paragraph{Personalized reports, sharing, and creativity.}\label{sec:sharing}
One of the final steps of the data analysis workflow is to create dashboards, reports, and presentations to communicate the results and insights to stakeholders. Current tools like PowerBI, Tableau, and Powerpoint require significant proficiency and development effort. Some ways AI systems can assist here are by facilitating the generation of dashboards and what-if analyses with minimal developer involvement~\cite{pandey2022medley, wu2023socrates, sultanum2021leveraging}.
AI systems also bring the opportunity to transform static elements, such as papers and documents, into interactive versions~\cite{10.1145/3290605.3300295, heer2023living}. Another example is Microsoft Office copilots that answer questions on the side about the document in an interactive manner. 

Furthermore, AI systems, particularly LLMs, can tailor the content and format of reports to suit various audiences and devices by dynamically adjusting tone, detail level, and emphasis.
Another aspect in which LLMs and other types of GenAI models such as DALL-E and Stable Diffusion can help is by creating artistic and aesthetically pleasing charts and infographics~\cite{dibia2023lida, schetinger2023doom}. By interpreting the underlying patterns and insights in a dataset, GenAI models can craft visually representative charts that not only communicate information effectively but also evoke a sense of creativity and design. For example, for a chart regarding global warming trends, we can create an infographic with a melting ice metaphor to convey the urgency and severity of climate change. 

In general, creating interactive and personalized artifacts for communication is challenging, as it involves linking various forms of media such as static text, images, videos, and animations, as well as employing diverse interaction techniques such as details-on-demand, drag-and-drop functionality, scroll-based interactions, hover effects, responsive design, and gamification to enrich communication~\cite{hohman2020communicating}. 
The vast space of potential interactive designs presents an interesting problem for AI-based systems, which could generate these artifacts by understanding user preferences and intentions and executing them accordingly.

\section{Human-driven design considerations for AI-based data analysis systems}\label{sec:design}
\begin{figure*}
\includegraphics[width=\linewidth]{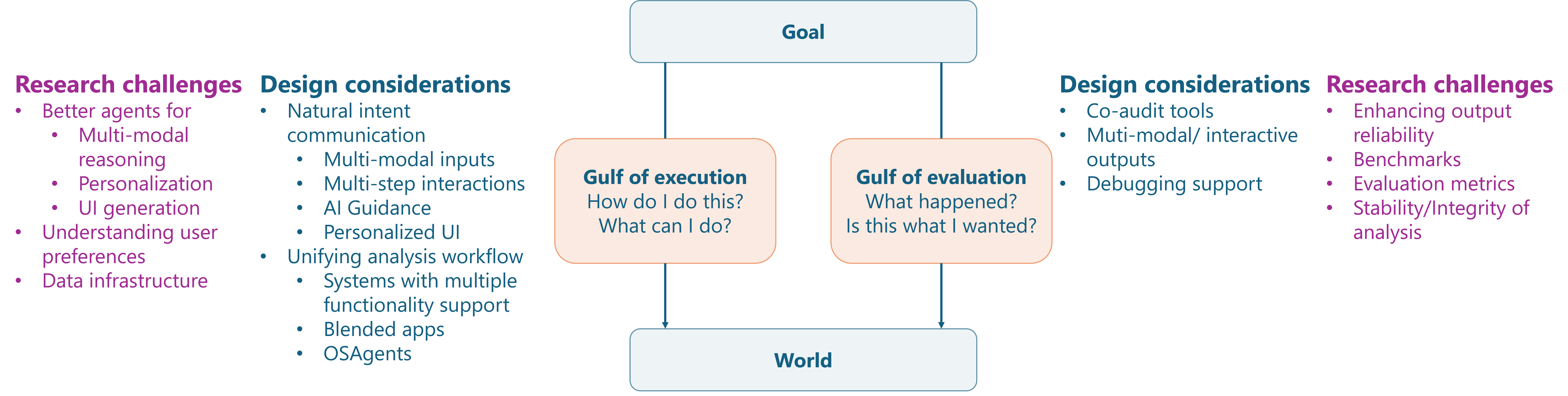}
\caption{Human-driven design principles and research challenges based on the gulf of execution and gulf of evaluation principles from~\cite{hutchins1985direct}. }
\label{fig:challenges}
\end{figure*}

In the previous section, we explored how AI systems can impact various workflows within the data analysis pipeline. However, the user experience of these AI-powered workflows varies significantly based on several design considerations. As Hutchins et al.,~\cite{hutchins1985direct} highlighted, two main challenges users must overcome when interacting with technology are (a) Execution: taking action to accomplish a particular goal, and (b) Evaluation: understanding the state of the system. While GenAI alters the notions of execution (for e.g., specifying intentions instead of writing visualization code) and evaluation (for e.g., understanding AI-generated code in addition to reading charts), the gulfs of execution and evaluation still exist. Therefore, it is crucial to design AI systems with users in mind, aiming to reduce these new gulfs.~\autoref{fig:challenges} presents an overview of potential  human-driven design considerations, as well as outlines the research challenges that need to be addressed to implement these designs, which will be discussed in more detail in~\autoref{sec:system-challenges}.

\subsection{Enhancing user-AI interaction: For natural intent communication}
Relying solely on natural language based interfaces can pose challenges for users of AI-based data analysis systems, as expressing complex intents can become lengthy or difficult due to limited expressiveness. For instance, altering the legend placement in a chart requires users to acquire the literacy to talk about charts (in this case, learn the word ``legend'') so that an AI agent would unambiguously understand. Similarly, for complex tasks, users might find it necessary/beneficial to interact with intermediate artifacts, explore variations, and integrate them gradually, rather than specifying everything at once. 
Below we discuss some potential ways AI systems can enable natural modes~\footnote{by natural modes, we mean straightforward (possibly personalized) modes that do not require users to learn new communication skills} for users to communicate their intent. 

\paragraph{Multi-modal inputs.} Beyond chat and natural language interaction, there are other modalities like mouse-based operations, pen and touch, GUI based interactions, audio, and gestures that could provide a more powerful and easier way for users to provide their intents. 

For example, Data Formulator~\cite{wang2023data} is an AI system that supports multi-modal inputs. Data Formulator combines graphical widgets and natural language inputs to let users specify their intent more clearly with less overhead compared to using just natural language interface (\autoref{fig:run_ex_io}(c)). Data Formulator provides a shelf-configuration user interface~\cite{grammel2013survey} for the user to specify visual encodings, and the user only needs to provide short descriptions to augment UI inputs to explain chart semantics. In Data Formulator, the shelf is pre-populated based on the current columns in the dataset, but one could also imagine auto-generating additional concepts on the fly based on the task and context and let users choose the right concepts rather than fully specifying them in natural language. 
DirectGPT~\cite{masson2024directgpt} is another system that allows users to specify parts of the intent by clicking, highlighting, and manipulating its canvas. This type of direct manipulation has been previously explored for image generation~\cite{dang2022ganslider}. InternGPT~\cite{liu2023internchat} combines chatbots like ChatGPT with non-verbal instructions like pointing movements that enable users to directly manipulate images or videos on the screen.

Multimodality can also surface as demonstrations from the users. Pure demonstration-based inputs have been explored before~\cite{barman2016ringer}, but combining demonstrations with natural language can be complex. For example, ONYX~\cite{ruoff2023onyx} is an intelligent agent that interactively learns to interpret new natural language utterances by combining natural language programming and programming-by-demonstration.
Sketching or inking is another way for users to communicate their intent. Especially for designing charts, users often have a clear idea of the chart design they want, which can be challenging to specify using natural language alone~\cite{lee2013sketchstory}.

Another form of user specification can be in the form of input-output examples. For instance, programming by example is a very common strategy to specify data transformations like pivoting~\cite{gulwani2016programming}. However, traditional non-AI approaches cannot handle noise (or manual errors) in the provided examples, which LLMs can potentially alleviate. 

Audio input can be viewed as another representation of chat, but it is more powerful when combined by demonstrations or other user manipulations by associating specific utterances to what the user was doing at that point of time. 

Supporting multimodal inputs requires careful system design. For example, we need to be able to automatically map user's interaction on a chart/UI visually to the appropriate underlying component~\cite{you2024ferret, liu2018data}. We might even require  entities (like charts) to be modular and interactive to enable actions like drag-and-drop. Dealing with continuous inputs like audio and real-time screen activity is also challenging for current models. However, recent model innovations such as GPT-4o~\cite{gpt4o}, which support audio, video, and text input forms, are promising. Some models such as InternGPT~\cite{liu2023internchat} also finetune large vision-language models for high-quality multi-modal dialogue.

\paragraph{Multi-step interactions. }
Human-AI interaction is often a multi-step process. Initial user specifications can be ambiguous, requiring adjustments based on outputs, or the tasks themselves might be multi-step, driven by feedback from intermediate results, such as in data exploration. Thus, enabling users to interact and iterate during the analysis process is crucial.

Some linear forms of interactions include chat based interaction, that aggregates context from previous conversations~\cite{wang2023chatcoder}, and computation notebook style, that enables users to inspect intermediate outputs. However, sometimes non-linear forms are needed, such as generating multiple outputs (charts, designs, etc.) in the first step and allowing users to choose or combine them to proceed further. Users might also need to backtrack and fork again. Threads and tree like representations can be used for non-linear interactions~\cite{dataformulator2,hong2023conversational}. For example, Data Formulator2~\cite{dataformulator2} organizes the user’s interaction history as data threads to help the user manage analysis sessions (\autoref{fig:run_ex_it}(c)). With data threads, the user can easily locate an existing plot to refine, fork a new branch from a previous step to explore alternatives, or backtrack to a previous state to correct mistakes.  AI Threads~\cite{hong2023conversational} is another system that facilitates multi-threaded conversations, enhancing a chatbot’s contextual focus and improving
conversational coherence. Sensescape~\cite{suh2023sensecape} provides multilevel abstraction and seamlessly switches between foraging and sensemaking modes. Graphologue~\cite{jiang2023graphologue}, an interactive system that converts text-based responses from LLMs into graphical diagrams to facilitate followup information-seeking and question-answering tasks.

Tracking an AI agent's state and users' specifications along a particular path in the interaction tree is crucial for drafting the appropriate context for the models. Prompt summarization (such as~\cite{wang2023recursively}) is a potential technique to make sure that each node in the tree is stateless and appropriately captures the full context that led to that particular node.  Along with this, it is also necessary to have an appropriate user interface to enable users to interact in a potentially non-linear fashion with the AI agents. 

\paragraph{AI guidance vs user control.}
Another important aspect of user-AI interaction design is whether users start with a blank slate, warm start with AI-driven recommendations, or use a mixed-initiative approach where both the AI system and the user take the initiative as appropriate. Starting with a blank state empowers users with maximum flexibility and control over their tasks, whereas AI-driven recommendations offer proactive guidance and suggestions, especially when users are unsure of what to do next or unsure of what the AI system can do for them. For example, Lida~\cite{dibia2023lida} exemplifies this approach by presenting users with a list of potential goals upon loading a dataset as well as recommendations of relevant visualizations once an initial visualization has been created. Similarly, various copilots and tool assistants  offer prompt suggestions to guide user interactions.
Some AI assistant systems can also prepopulate entire documents or slides based on notes or other user documents, providing users with a starting point that they can edit and refine through follow-up interactions.

GenAI models are increasingly gaining the ability to analyze datasets, user history, and the behaviors of other users to predict and recommend next steps, or even entire workflows. This shifts user interaction from building components from scratch to refining AI-generated suggestions. For example, after a model generates a slide deck, users might probe the system to understand/modify the data that was used to create the charts in the slide. This shift underscores the need for AI systems to support probing mechanisms, offering interfaces that allow users to ask follow-up questions, intervene and modify specific parts, and provide clear explanations of the AI's processes and decisions.

\paragraph{Personalized and dynamic UI.}
The language model's generative capabilities present an opportunity to generate customized UIs on the fly. DynaVis~\cite{vaithilingam2024dynavis} is one such system that showcases LLM's capacity to dynamically generate UI widgets for the subtask of visualization editing. As the user describes an editing task in natural language, such as ``change the legend position to left'', DynaVis performs the edit and synthesizes a persistent widget that the user can interact with to make further modifications (to try other legend positions and find the one that best fits their needs) (\autoref{fig:run_ex_it}(b)). Stylette~\cite{kim2022stylette} is another system that allows users to specify design goals in natural language and uses an LLM to infer relevant CSS properties. PromptInfuser~\cite{10.1145/3544549.3585628} explored using prompting in creating functional LLM-based UI prototypes. 
LLMs have also been used to generate feedback on UI designs~\cite{duan2023towards}.Expanding on the above techniques to empower LLMs in generating personalized user interfaces for data analysis tools holds significant promise for enhancing user experience with tools in the future.

\subsection{Facilitating trust and verification: Enhancing model output reliability for users}\label{sec:trust}
GenAI models can produce unintended outputs either because of model hallucinations or the ambiguity of the user input. This makes it essential for users to verify the accuracy of the models' outputs. However, the verification process should not impose a cognitive load comparable to or greater than that of a person performing the task by themselves. For example, verifying data transformations may require more than visual inspection, requiring code review or editing.  Hence, it is important for AI system designers to also facilitate easier ways to allow users to verify the outputs. This aligns with the argument that GenAI systems impose significant metacognitive demands on users, as highlighted in~\cite{tankelevitch2024metacognitive}, where the authors propose integrating metacognitive support strategies into GenAI systems to reduce these demands, particularly by enhancing explainability and customizability. 

\paragraph{Co-Audit tools.}
There are several ways in which users can verify the quality of a model's output, such as gauging the high-level correctness from charts and transformed data, inspecting corner cases in data, understanding code, and conducting provenance analysis for selected data points~\cite{gu2024analysts}. AI agents can be designed to support these types of quality control, as highlighted by the emerging concept of co-audit tools~\cite{gordon2023co}, which aim to assist users in checking AI-generated content. 

One of the co-audit approaches is to provide code explanations either in natural language~\cite{wang2023data,cui2022codeexp} or ground the output code in editable step-by-step natural language utterances~\cite{liu2023wants}. 
Another verification approach is to generate multiple possible charts/data for the same user instruction and show any differences to the user to disambiguate~\cite{mundler2023self, lahiri2022interactive} (\autoref{fig:run_ex_trust}(b)). Inspection tools, such as ColDeco~\cite{ferdowsi2023coldeco}, provide spreadsheet interfaces to inspect and verify calculated columns without requiring users to view the underlying code. Provenance analysis is another critical aspect that tracks how a particular output (for e.g. a single data point) is obtained through a series of transformations from the input. \cite{buneman2001and} provides one of the early characterization of data provenance especially for data obtained through SQL queries. There are several recent systems such as XNLI~\cite{feng2023xnli}, which shows the detailed visual transformation process to the user along with a suite of interactive widgets to support error adjustments and a mechanism for the user to provide hints for revision in the data visualization domain. Datamations~\cite{pu2021datamations} is another tool that shows animated explanation of the various steps that led to a particular plot or table. 

\paragraph{Multi-modal and interactive outputs.}
Additionally, the format of the output has a significant influence on the user's understanding and verification ability. For example, a long text describing the analysis results and insights can be difficult for users to parse, and lengthy code can be challenging for users to understand. But a multimodal document interlaced with text and images will be more readable.  Similarly, interactive charts and what-if analyses allow users to quickly experiment with different parameters, helping them verify outcomes more effectively than by reading through extensive code. Dynamic user interfaces customized for the user and the task in the style of DynaVis~\cite{vaithilingam2024dynavis} can also help by providing UI and widgets on the fly to help users interactively probe and refine AI generated outputs. 

\paragraph{Debugging support for users.}
AI systems, which often consist of multiple models, agents, and components like retrieval-augmented generation (RAG) or self-repair mechanisms, can be challenging for end users to debug or understand. Users typically see only the final outcome without insight into the internal workings of the system, making it difficult to grasp why the AI system is behaving in a certain way, producing specific outputs, or failing to complete tasks.

To improve trust, one potential strategy is to design these systems to provide real-time visibility into the actions of these agents, along with support for checkpointing and interruptibility. This would allow users to stop the process at any given point, provide feedback, and then resume.
Recent work, such as AutoGen Studio \cite{dibia2024autogenstudio} which supports building of AI systems by orchestrating multiple agents to collaborate, addresses this challenge by providing a no-code interface tool for authoring, debugging, and deploying multi-agent workflows.  
In particular, AutoGen Studio's debugging view  visualizes the messages exchanged between agents using charts and displays key metrics such as costs (tokens used by an LLM, dollar costs, number of agent turns, and duration), tool usage, and tool execution status (e.g., failed code execution).

\subsection{Unified analysis experience: Streamlining data tools and workflows} \label{sec:multi-app} While most current data analysis tools focus on automating one or a few steps of the data analysis pipeline, upcoming innovations in AI models and complex multi-agent AI systems have the potential to change the entire end-to-end data analysis stack. 
In the current workflow, users must navigate multiple tools like Excel, PowerBI, Jupyter Notebooks, and SQL servers because each serves a unique purpose and offers interfaces tailored to different aspects of data analysis. Even with GenAI techniques integrated into these tools, significant overhead remains in switching contexts, transferring data, and carrying over intent from one application to another. Therefore, another key design consideration for AI-powered data analysis tools is finding ways to streamline this process for users.

\paragraph{One tool multiple capabilities.}
Unifying multiple capabilities into a single tool is one  way to streamline the analysis process by allowing users to work in a single environment, eliminating the need to re-explain domain contexts and provide feedback across multiple platforms. 
It also  enables the unified AI system to seamlessly debug and reason across the various steps of the process.  For example, if a visualization appears inaccurate due to improper data cleaning, the unified AI system can identify and address issues, without user intervention to identify and debug the problem.

Some of the AI-powered tools we have mentioned so far try to combine more than one data analysis step into a single tool. For example, Data Formulator~\cite{wang2023data} unifies data transformation with visualization authoring, which improves the user experience over traditional visualization tools (e.g., PowerBI, Tableau) that need users to perform a separate step to transform their data to tidy formats as needed by the respective tools. LIDA~\cite{dibia2023lida} is another tool that combines data summarization, goal exploration, chart authoring, and infographics generation into one tool. 

However, it may not always be feasible to design a single tool that supports all stages of data analysis and accommodates the diverse preferences of users, which motivates alternate approaches, such as blended apps and OSAgents, which are discussed below.

\paragraph{Multi-agent systems.}
A key advancement that enables the integration of multiple capabilities within a single tool is the development of multi-agent systems~\cite{wu2023autogen, hong2023metagpt}. These systems allow us to create AI systems composed of multiple agents, each specialized in a particular capability, and configure the agents to interact with one another to plan and collaborate on complex tasks. They have shown promise in producing complex AI tools in various domains such as  scientific reasoning~\cite{lu2024chameleon}, software engineering~\cite{hong2023metagpt, qian2023communicative}, and embodied agents~\cite{gong2023mindagent}. Closest to the data analysis domain is the software engineering domain, where systems such as ChatDev~\cite{qian2023communicative} are modeled as a system of AI ``software engineers'' such as programmers, code reviewers, designers, and testers who collaborate to facilitate a seamless software engineering workflow. They show that this kind of architecture in conjunction with high quality models is effective  in identifying and alleviating potential software bugs, and also at reducing hallucinations of the LLMs. 

\paragraph{Blended apps.}
Modern applications such as Microsoft Loop and Notion are starting  to create all-in-one workbenches, where one brings multiple tools/data that are needed for the task into a single interface. 
In the context of building AI tools for the data analysis domain, a similar approach would be  to design AI tools that can be contextualized in the overall analysis process. Individual AI systems for one part of the analysis workflow should be able to talk to AI systems for the other parts, share their contexts, and collectively solve/debug tasks. Compared to a single AI system handling multiple capabilities, this approach allows developers to build AI systems for different capabilities separately, each with its own customized user interface and verification strategies, while still providing users with a unified experience.
Additionally, the user experience can be further improved by AI systems talking to and extracting context from existing non-AI systems. For example, going back to the Data Formulator tool, even though it helps users automate certain data transformations like table pivoting, it does not allow users to perform manual cleaning of the data; however, users can easily manually edit data in tools like Microsoft Excel. Therefore, a blended app experience that transitions automatically and ports any necessary context from one tool (AI or non-AI based) to another can enhance the user experience drastically.

\paragraph{OS Agents.}
Another notable way to reduce tool overhead is by using ``OS Agents'' or a ``Self-driving OS'' that automatically controls and interacts with multiple applications on the desktop. For example, UFO~\cite{zhang2024ufo} is a GPT vision-based agent that observes and analyzes the GUI and control information on Windows applications and performs tasks across multiple applications based on  natural language commands from the user (\autoref{fig:run_ex_apps}(b)). For applications that cannot be controlled via their APIs or co-pilots, this way of direct UI manipulation with AI agents is a promising means to enable a unified analysis experience for end-users.

\section{Challenges in developing AI powered data analysis systems}
\label{sec:system-challenges}
Now that we have explored various design considerations to enhance human-AI experiences in data analysis tools, several research challenges remain in developing AI-powered systems that effectively incorporate these design principles.
As outlined in \autoref{fig:challenges}, these include developing new models to support multimodality, planning, and personalization; understanding user preferences; creating data infrastructure to facilitate AI-driven suggestions; enhancing the reliability of existing AI models; and establishing robust system benchmarking and evaluation metrics. While some of these challenges are common across any LLM-based interactive system~\cite{kaddour2023challenges}, this section  particularly contextualizes the problems to the data analysis domain. 
\subsection{Ensuring reliability and trust}
AI models are known to suffer from several issues such as hallucinations~\cite{tonmoy2024comprehensive}, sensitivity to prompts~\cite{sclar2023quantifying}, failure to follow instructions~\cite{liu2024lost}, lack of acknowledgment of uncertainty~\cite{key2022toward}, and biases~\cite{gallegos2023bias}. This is true even for the data analysis domain~\cite{liu2023wants, singha2023tabular}. 
For e.g. ~\cite{singha2023tabular} shows how LLMs' performance is suspectible to table representation in the prompts.  
Thus, given the potential consequences of bad results, it is crucial to improve the reliability of AI-based data analysis systems. 
\paragraph{Improving code correctness of LLMs.} The first step is to ensure that LLM's output satisfies the specification, such as using the correct API and being syntax error free.
Grounding techniques, such as RAG (Retrieval-Augmented Generation) and GPT-4 Assistants API, enable LLMs to ground their outputs in external knowledge such as API documentation and relevant examples, as well as use tools such as calculator and custom functions to reduce errors in the output~\cite{zhang2023toolcoder, patil2023gorilla,peng2023check}. For code generation parts of the data analysis pipeline, we can increase the accuracy of the models by verifying the code against a suite of input-output examples (which are either provided by the user~\cite{wang2023data, wen2024grounding} or automatically generated but verified by the user~\cite{lahiri2022interactive}). It has also been shown that
self-repair~\cite{gero2023self, olausson2023self} and self-rank~\cite{inala2022fault} mechanisms that let models evaluate their own outputs, and even repair them, have been very beneficial in improving the accuracy of the code generated by the models. 

\paragraph{Handling failure cases.}
Sometimes, despite the above efforts, models might still produce undesirable outcomes (such as a code that crashes). Some strategies for handling such failures include providing fallback options or having the model intelligently request additional information from the user. Consider generating UIs on the fly using LLMs, where the robustness of the output is crucial to maintain a seamless user experience. In addition to having mechanisms for error identification and correction, incorporating fallback options such as static UIs or dynamically composing from static UI elements can mitigate disruptions caused by the errors.

\paragraph{Ensuring Stability and Integrity of the Analysis.}

Many factors, such as ambiguous user intents, faulty data, model biases, and hallucinations, can cause an AI system to produce unintended data analysis outputs. In the analysis process, LLMs may be used for hypothesis generation, making assumptions about the dataset or domain, or generating statistical reasoning. Therefore, it is essential to ensure that these assumptions and hypotheses are not flawed due to the above factors. Therefore, AI system developers should assess the stability and integrity of the AI system's outputs before presenting them to users. 

To identify hallucinations and  ambiguous user intents, one approach involves sampling multiple outputs from the LLMs and examining their agreement to detect inconsistencies. 
The predictability, computability, and stability (PCS) framework~\cite{yu2020veridical} offers another structured method for evaluating the trustworthiness of results by ensuring that they are predictable, stable, and aligned with real-world contexts. A key aspect of this framework is demonstrating that the analysis outcome remains consistent when slight perturbations are applied to inputs, assumptions, and hypotheses at various stages of the data analysis pipeline.

Another threat to AI-driven analysis is the possibility of overlooking crucial considerations during hypothesis exploration, even if the analysis is stable. For example, when selecting a store with strong sales performance, the analysis might focus solely on total profit but miss important factors such as the store's profitability relative to its size or operational costs, which could lead to misleading conclusions. One potential approach to identify such failures is to actively probe the model to consider alternative analyses  (for e.g., for the above scenario, the AI system can actively probe the models to output an analysis that makes a different store than what was predicted before to appear as the strong performer). After generating multiple possible analyses, the system can reason over them to determine which makes the most sense given the user’s intents, the task, and the domain.

Overall, ensuring the integrity of the entire AI-driven analysis workflow is significantly more challenging than ensuring the reliability of individual AI steps, such as code generation for charts.  This requires developing new frameworks, such as the PCS framework, and incorporating active probing, mutating, and testing strategies as integral components within the AI-data analysis system.

\subsection{System benchmarking and evaluation metrics} 

\paragraph{Desired evaluation benchmarks.}
To enable researchers to prototype and evaluate data analysis models and systems effectively, we need a comprehensive benchmark suite that covers a broad range of data sources and tasks within the data analysis domain. This suite should include both low-level tasks for each step of the data analysis process and high-level tasks that span multiple steps. Other domains, such as AI for Code, already have rich and rapidly growing benchmark sets (e.g., HumanEval~\cite{chen2021evaluating}, CodeContests~\cite{li2022competition}, and cross file code completion tasks~\cite{ding2024crosscodeeval}) that helped evaluate new techniques, such as self-repair~\cite{olausson2023self} and test-driven generation~\cite{lahiri2022interactive}. The goal is to create a similar resource in the data analysis domain.

While the data analysis domain already has several benchmark collections, they often focus on specific steps of the data analysis pipeline. For example, existing benchmarks address visualization authoring~\cite{luo2021nvbench, srinivasan2021collecting}, data transformations (DS1000~\cite{lai2023ds}, Juice~\cite{agashe2019juice}, CoCoNote~\cite{huang2024contextualized}), machine learning (MLAgentBench~\cite{huang2024mlagentbench}), and statistical reasoning (Blade~\cite{gu2024blade}, QRData~\cite{liu2024llms}).  DABench~\cite{wang2024dabench}, another benchmark, tests LLMs' abilities to write and execute complex code for tasks such as outlier detection and correlation analysis, which are essential for hypothesis evaluation.  These benchmarks are very valuable, but they are spread across multiple benchmark suites, each with different evaluation metrics and procedures devised by individual research groups. Moreover, they often use data sources and tasks of varying difficulty levels, making it challenging to compare AI systems across different suites. Additionally, most of these benchmarks focus on the code generation and the code execution aspect of the data analysis pipeline and have tasks with unambiguous intents and single word, easily evaluated answers, which limits their ability to test all aspects of AI systems in the data analysis domain.

Moreover, benchmarks are often created to highlight specific model capabilities or techniques, typically tailored to the strengths of existing models. This approach can result in narrow benchmarks that do not fully represent the complexities found in real-world data analysis. Therefore, more research is needed to develop benchmarks that reflect the diverse challenges encountered by data analysts, such as tasks requiring multiple iterations of analysis, gathering additional data based on previous results, fixing data issues before creating visualizations, and exploring hypotheses. These domain-oriented benchmarks can also reveal the limitations of current models in areas such as multi-modal reasoning, multi-step planning, and exploration versus generation capabilities of models, driving AI researchers to push the boundaries of their models and techniques in new directions.
Therefore, it would be useful to compile a diverse set of data tables and sources into a centralized location, along with potential high-level questions or goals (e.g., decision-making, report creation) and low-level concrete tasks for each step of the data analysis process. A taxonomy of tasks, categorized by complexity for both humans and models (e.g., tasks requiring multimodal reasoning, planning, or exploration), would help researchers assess test coverage, understand the strengths and limitations of AI systems, and identify areas for improvement.

There are  other interesting kinds of benchmarks that  are still lacking in the community. These include benchmark tasks that involve potential human intervention, where the full intent isn't known at the outset and evolves based on what the model generates in earlier steps. We also need benchmarks that require models to engage in open-ended exploration and iterative reasoning, similar to human scientific reasoning. Additionally, it would be beneficial to develop session-based tasks that capture a user's activity during an entire session or across multiple sessions. These tasks could involve scenarios that would demand  multi-modal and dynamic UIs, such as when users need to specify something difficult to articulate in natural language, repeat certain actions, or analyze multiple datasets in a single session. Such benchmarks would also help evaluate an AI system's ability to personalize to user preferences over time, track context based on past actions, and reuse contexts from previous sessions.

\paragraph{Evaluation metric challenges.}
Benchmarks should be accompanied by robust evaluation metrics. While it is valuable to evaluate a system in real usage with users in their own workflows with user studies and telemetry data, this process is often expensive and hence, are only used after promising results are observed in controlled offline environment.  
 These Offline metrics are useful for fast prototyping because they do not require real time user interaction. 
 
 For general purpose coding tasks, there are some established ways to measure accuracy of the output by adding a suite of test cases (input and expected output) for each task and executing the generated code on these test cases to check the outputs match the expected output in the test case. However, for the data analysis domain, the generated code executes to produce complex artifacts such as charts or UIs which are hard to compare because they are subjective. Moreover the design principle of supporting  multi-modal forms of communication (i.e., natural language, gestures, etc.) and interactive  and iterative human-AI collaboration, makes offline evaluation even more challenging.  To address these challenges, there are some preliminary techniques for using a model such as GPT4, GPT-Vision to simulate a human in the loop to both judge and evaluate multi-modal artifacts~\cite{liu2023gpteval} and to provide intermediate inputs and interact with GUIs by generating web inputs~\cite{liu2023chatting}.

Another key consideration in designing evaluation metrics is the ability to assess partial correctness of AI systems. For example, an AI system might successfully explore multiple hypotheses but make a small mistake in one, leading to cascading effects on the final output. Additionally, evaluation metrics should account for multiple orthogonal dimensions of performance. For example, an AI system might excel at task breakdown, general code generation but struggle  using particular visualization APIs or creating intuitive designs, or vice versa.

\paragraph{Behavioral evaluation.}
 In addition to assessing the correctness of AI-generated outputs for data analysis tasks—i.e., whether the output matches the expected result--it is equally important to evaluate whether the AI system’s output fosters user trust. For instance, when asked which car is the best to buy, a system that provides an answer along with explanations of why the car is better across multiple dimensions, and details how it performed calculations as part of the analysis, is more trustworthy than a system that simply gives an answer, even if the answer is the same. Similarly, it's important to evaluate not only whether the model can, on average, produce the correct answer (such as the pass@k metric, which assesses if the model provides a correct answer within k attempts), but also measure worst-case metrics such as how frequently the model hallucinates, as wrong answers have potentially bad consequences. Lastly, it is crucial to include tasks and evaluation metrics that explicitly perform adversarial testing of the AI system. (for e.g. variations of prompts that can trigger a misleading but convincing analyses), similar to the adversarial robustness experiments conducted for NLP tasks in~\cite{raina2024llm, kumar2023certifying}.

\subsection{Need more advances in models/agents}
We need further advances in models and agents to fully realize multi-modal, iterative, and trustworthy AI systems for the data analysis domain. 

\paragraph{Inference cost.} Firstly, while LLMs like GPT-4 offer powerful capabilities, they come with high inference costs. Conversely, smaller models such as Llama and Phi-3 may not consistently produce reliable outputs. Hence, there is a need for advancements in smaller language models to strike a balance between efficiency and accuracy.

\paragraph{Training data.}
One of the challenges of using LLMs for data analysis is that many of the latest models are not specifically trained on data-related or data-analysis-specific tasks.  While there may be adequate data coverage for programming languages such as Python or JavaScript, there is a lack of training data for other languages used in data analysis, such as R or VBA scripts. Additionally, although LLMs are potentially trained on web content and UI designs, they lack sufficient training data on UI interactions, such as capturing what happens when a user clicks a specific button in applications like Excel. This kind of data is crucial for LLMs to generate personalized UIs to help user better interact with the AI system. Another challenge arises from the difficulty of finding training data that maps high-level user intents to low-level data analysis tasks, since the underlying human intent (such as why a particular analysis is performed) is often not apparent in the final artifacts--such as Jupyter notebook code--which are scraped from the web for training models.

\paragraph{Finetuning; RLAIF; Multi-agent finetuning.}
Finetuning the AI models for specific data-analysis capabilities on a smaller dataset is one possible solution to overcome the lack of training data problem. Avenues for finetuning in this domain include finetuning pre-trained LLMs to understand the semantics of the tables/data~\cite{li2023table, zhang2023tablellama, dong2022table},  finetuning to learn to generate domain-specific code/actions for which there is not much training data, and finetuning multi-agent systems over the entire data-analysis workflow. A significant challenge in fine-tuning is the scarcity of ground truth data. For multi-agent workflows, for instance, there is no supervised data to guide the intermediate actions of each agent. Similarly, in newer code domains or user-interaction-based tasks, it is difficult to gather annotated data manually. Hence, automated ways of gathering ground truth data are useful. Here, we can leverage the same techniques that are needed for automatic offline-evaluation of AI-data analysis systems; utilizing current AI models to generate multiple candidate outputs and then employing offline evaluation to label these candidates. This technique is called RLAIF~\cite{lee2023rlaif} as it  involves using AI feedback, such as simulated human agents, to enhance models  (especially to improve smaller models using feedback from larger models).

\paragraph{Personalization and continual learning}
Personalizing AI system's outputs is crucial for enhancing user experience. As users are likely to engage with AI-based data analysis systems across multiple sessions, AI agents need to learn from previous interactions to understand user preferences. Advancements in agents and AI models are needed to enable memory storage~\cite{wang2024augmenting}, learning from past actions and human feedback, and adapting to user preferences over time (for example, the amount of code they would like to see, colors they would like in their plots, their reporting style, etc.). Additionally, AI agents should be able to leverage data from other users, offer tailored recommendations, and predict users' next steps similar to current recommendation systems  used by streaming platforms and e-commerce websites.

Lastly,  AI systems that can evolve and adapt to new tasks over time is an exciting area of research. For instance, if an AI system is repeatedly used to perform actions on an Excel document via the OfficeJS API,  it  could potentially learn to improve its use of the existing API. Moreover,  it could also learn to create new APIs and better abstractions that are optimized for the most common user queries. For example, let's say the original API lacks direct support for certain tasks, for which the AI agent may need to write longer, more complex code, increasing the chances of failure. In this case, can the AI system through its self-evolution process create a more specialized, direct API to make the generation process   more reliable?

\paragraph{Multi-modal reasoning.}
The data analysis tasks require AI models to reason across multi-modalities such as code, text, speech, gestures, images, and table modalities. As multi-modal models such as GPT-4o, Llava, Phi-3-Vision evolve, we can expect improvements in AI-assisted data analysis tools. However, current models face several challenges: for example, LMMs are often trained on natural images (e.g., scenery, human faces) and struggle to understand images such as charts~\cite{wu2024evaluating}.
Understanding gestures is another challenge, for instance, drawing an arrow between two columns in a table might indicate that the user wants to add a new column at that location. Additionally, while current models are not yet perfect in any single modality (e.g., interpreting a chart from the image modality), they can improve performance by combining reasoning across multiple modalities, such as analyzing a chart, the underlying data, and the output of code executions simultaneously.

\paragraph{Planning and exploration.} Data analysis is not a linear process; it involves planning, reasoning, and exploration, requiring AI agents to be skilled at hierarchical planning, iterative problem solving, and flexible adaptation to new information and unexpected challenges. LLM-based planning has been explored in other contexts, such as embodied agents~\cite{xi2023rise}. However, evaluations on international planning competition problems have yielded mixed results~\cite{valmeekam2023planning}, showing that LLMs perform better when collaborating with humans rather than functioning autonomously. This highlights the need for further advancements in both LLM capabilities and human-AI collaboration for effective planning for data-analysis tasks.

\subsection{Understanding user preferences and abilities}
A key component to building interactive AI systems is understanding user preferences and abilities to tailor an experience that is smooth and intuitive for the users. This can be in the form of formative studies to inform and shape the design process such as~\cite{gu2024analysts, mcnutt2023design} or summative studies to assess the performance of a system after development such as~\cite{wang2023data, vaithilingam2024dynavis}. Understanding user preferences can not only guide system design but  can also potentially be integrated into the AI system itself, allowing it to deduce user preferences based on user's actions, history, and the task as the session progresses. 

Existing user studies have provided valuable insights into user preferences in AI-driven data analysis. On user interaction preferences, the study from Data Formulator2 has found that the GUI + NL approach  is more effective than chat-based AI assistants  in communicating and constraining intent. Another study on the DynaVis tool~\cite{vaithilingam2024dynavis} revealed that participants  preferred AI-generated widgets that they can use to perform actions rather than having the AI perform actions directly, However, they also noted that frequently changing UIs could increase cognitive load. On preferences on AI suggestions and outputs,~\cite{gu2024data} found that users' varying levels of statistical expertise influenced their reactions to AI-generated suggestions, with some finding them helpful and well-matched to their expertise, while others found them overly basic or requiring significant effort to understand due to mismatches with their background. In another study on how users verify AI-assisted data analysis~\cite{gu2024analysts}, participants needed  both procedure-oriented artifacts (e.g., natural language explanations, code, and code comments) for understanding analysis steps and data artifacts (e.g., intermediate/output data tables and summary visualizations) for confirming low level details. Finally, on users' preference on understanding and leveraging AI capabilities, the study performed by~\cite{mcnutt2023design} has found that users want to understand and control the context given to the LLM model to ensure relevant and accurate outputs and users like linter-like assistants that highlight inappropriate usage of AI systems for users. 

Despite these insights, there are still gaps in our understanding of user preferences.For example, how can we deduce contextual preferences, such as the most effective modalities or the desired level of AI autonomy, based on user history, past actions, and evolving tasks?. There’s also a gap in understanding how to personalize UI elements for individual users without overwhelming them with frequent changes. Furthermore, research is needed on dynamic multi-application environments and AI-coordinated complex multi-step workflows, focusing, for example, on user intervention interfaces, user's ability to understand/manage AI-context  across multiple apps, UI designs that span different applications, and trust and verification with  output artifacts that might span multiple apps.

\subsection{Data Infrastructure}
Another key challenge is  ensuring the availability of high-quality data tables in various domains that AI systems can use to provide initiative analysis suggestions for users. 
This infrastructure might involve search engine style infrastructure such as crawling, indexing, and ranking data tables from the internet so that the AI system can easily extract the data tables necessary for user queries.  We also need infrastructure to enable domain experts to create data table APIs, facilitate real-time updates of the data tables using online data, and manage enterprise and proprietary data tables effectively. Additionally, mechanisms for evaluating and ranking data tables for quality, potentially through crowdsourcing, would further enhance the reliability and relevance of available data resources. Another aspect to consider is data privacy, security, and having mechanisms to anonymize and aggregate data, which is a significant research area in itself, but beyond the scope of this work.

\section{Conclusion}
This paper explores the vast potential that  generative AI-powered tools have in unlocking actionable insights from data. We first outline a set of tasks that AI systems can assist with, providing related works for some tasks and inspiration for others that are ripe for further exploration.
We then discuss human-driven design considerations aimed at optimizing user interactions, workflows, and enhancing trust and reliability. Most of these design considerations can in fact be applied more broadly to enhance the design and functionality of AI systems in other domains.
Finally, our work highlights research challenges such as improving the robustness of AI models in data analysis scenarios and developing benchmarks and evaluation metrics. We also note  several model advancements that are needed such as improved continual learning, multi-modal reasoning, and planning and that we can leverage multi-step, multi-modal interactive data analysis scenarios as benchmarks to drive these model innovations.
In addition, we emphasize the need for comprehensive user studies to discern preferences and optimize AI-driven tools for data analysis users. 
Our goal is that these efforts collectively will help bridge the gap between complex data and actionable insights for a wide-range of users.

\FloatBarrier
{
    \footnotesize
    %\scriptsize
    % \setstretch{1.0}
    \bibliography{main.bib}
    \bibliographystyle{icml2024}
}

\end{document}